\pdfoutput=1

\documentclass[11pt]{article}

\usepackage{ACL2023}

\usepackage{adjustbox}
\usepackage{times}
\usepackage{latexsym}
\usepackage{graphicx}
\usepackage{graphics}
\usepackage{CJK}
\usepackage{tabularx}
\usepackage{bm}
\usepackage{amssymb}
\usepackage{booktabs}
\usepackage{url}
\usepackage{multicol}
\usepackage{tabularx}
\usepackage{booktabs}
\usepackage{color}
\usepackage{arydshln}
\usepackage{multirow}
\usepackage{amsmath} 
\usepackage{bm}
\usepackage{subfigure}
\usepackage{amstext}
\usepackage{tikz}
\usepackage{pgfplots}
\usepackage{hyperref}
\usepackage{subfigure}
\usepackage{enumitem}
\usepackage{mathtools}
\usepackage{blindtext}
\usepackage{makecell}

\usepackage[T1]{fontenc}

\usepackage[utf8]{inputenc}

\usepackage{microtype}

\usepackage{inconsolata}

\usepackage{hyperref}

\usepackage{graphicx}
\usepackage{epigraph}

\usepackage{comment}

\definecolor{byzantine}{rgb}{0.74,0.2,0.64}

\newcommand{\heart}{\ensuremath\heartsuit}
\newcommand{\diamondsmall}{\ensuremath\diamondsuit}

\title{\textit{Let's Negotiate!} A Survey of Negotiation Dialogue Systems}

\author{Haolan Zhan\textsuperscript{\rm \heart}, Yufei Wang\textsuperscript{\rm \heart},  Zhuang Li\textsuperscript{\rm \heart}, Tao Feng\textsuperscript{\rm \heart}, Yuncheng Hua\textsuperscript{\rm \heart}, Suraj Sharma\textsuperscript{\rm \diamondsmall}, \\ 
\textbf{Lizhen Qu}\textsuperscript{\rm \heart},  \textbf{Zhaleh Semnani Azad}\textsuperscript{\rm \diamondsmall}, \textbf{Ingrid Zukerman}\textsuperscript{\rm \heart}, \textbf{Gholamreza Haffari}\textsuperscript{\rm \heart} \\
\textsuperscript{\rm \heart} Department of Data Science \& AI, Monash University, Australia\\
\textsuperscript{\rm \diamondsmall}  California State University, Northridge, CA \\
\{firstname.lastname\}@monash.edu, \{suraj.sharma, zhaleh.semnaniazad\}@csun.edu\\ 
}
\pgfplotsset{compat=1.18} 
\begin{document}
\maketitle

\begin{abstract}

Negotiation is a crucial ability in human communication. Recently, there has been a resurgent  research
interest in negotiation dialogue systems, whose goal is to create intelligent agents that can assist people in resolving conflicts or reaching agreements. Although there have been many explorations
into negotiation dialogue systems, a systematic review of this task has not been performed to date. We aim to
fill this gap by investigating recent studies in the field of negotiation dialogue systems, and covering benchmarks,
evaluations and methodologies within the literature. We also discuss potential future directions, including multi-modal, multi-party
and cross-cultural negotiation scenarios. Our goal is to provide the community with a systematic overview of
negotiation dialogue systems and to inspire future research.

\end{abstract}

\section{Introduction}

Negotiation involves two or more individuals discussing goals and tactics to resolve conflicts, achieve mutual benefit, or find mutually acceptable solutions~\cite{fershtman1990importance,bazerman1993negotiating,lewicki2011essentials}. 
It is commonly used to manage conflict and is the primary give-and-take process by which people try to reach an agreement~\cite{fisher2011getting,lewicki2011essentials}. Negotiations can be cooperative or competitive and are used in various social settings such as informal, peer to peer, organizational, and diplomatic country to country settings~\cite{basave2016study} and thus the implications for enhancing outcomes are vast. However, humans are naturally subject to various biases and can be swayed by emotion during negotiations, making them inclined to overlook useful implicit information from other participants in the negotiation process and hindering optimal outcomes. Negotiators also often lack the necessary skills, training and knowledge to achieve their desired goals~\cite{walton1991behavioral}.

\begin{figure}
    \centering  
    \includegraphics[width=0.49\textwidth]{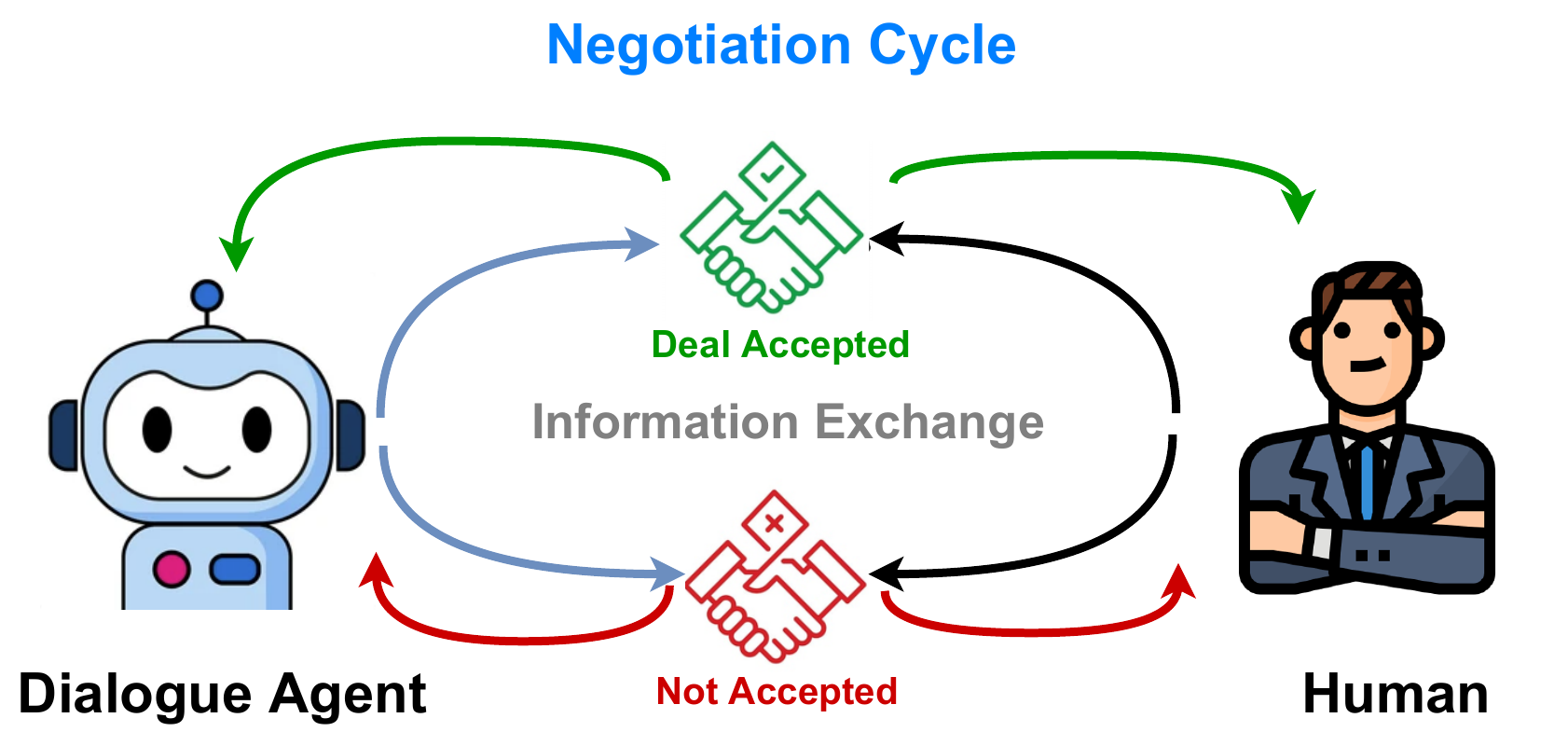}
    \caption{A typical negotiation dialogue involves a multi-turn interaction between agent and human. They exchange information about their deals and end up with accepting or declining deals.}
    \label{fig:intro}
\end{figure}

To facilitate human negotiation processes, previous researchers~\cite{lewandowska1982meaning,lambert1992modeling,chawla2021casino} have aimed to build intelligent negotiation agents that can aid humans or even directly negotiate with humans in multi-turn interactions (Figure~\ref{fig:intro}).
Effective agents could yield significant benefits in many real-world scenarios, ranging from bargaining prices in everyday life~\cite{he2018decoupling} to higher-stakes political or legal situations~\cite{basave2016study}.

Research on negotiation has been conducted for almost 60 years in the field of psychology, political science, and communication. 
It has evolved over the past decades from exploring game theory~\cite{walton1991behavioral}, behavior decisions driven by the cognitive revolution in psychology~\cite{bazerman1993negotiating}, to cultural differences in the 2000s~\cite{bazerman2000negotiation}. 
Negotiation research, however, is now forced to confront the implications of human/AI collaborations given recent advancements in machine learning~\citep{bazerman2000negotiation,ouali2017computational}.  
Research has focused on establishing new benchmarks and testing environments for various negotiation dialogue scenarios, including product price bargaining~\cite{lewis2017deal,heddaya2023language}, multiple player strategic games~\cite{stac2016discourse} and job interviews~\cite{zhou-etal-2019-dynamic}. 
Other research has attempted to propose new methodologies and frameworks to model the negotiation process, including various negotiation policy learning, negotiator mental status modeling and negotiation decision making.
Converging efforts from social scientists and data scientists which incorporate insights from both fields will thus be fruitful in maximizing processes and outcomes in negotiations.

Despite the significant amount of research that has been conducted, we are not aware of a systematic review on the topic. In this work, we aim to fill this gap by reviewing contemporary research efforts in the field of negotiation dialogue systems from the dimensions of datasets, evaluation metrics and modeling approaches. We first briefly explore human negotiations and corresponding limitations, and propose how dialogue agents may supplement human negotiation processes. We then discuss the popular negotiation dialogue modeling methods, including \emph{Strategy modeling}, \emph{Negotiator modeling} and \emph{Action modeling}. We further introduce existing datasets according to their negotiation scenarios. Finally, we give an overview for three major types of evaluation metrics, i.e., \emph{goal-based metrics}, \emph{game-based metrics} and \emph{human evaluation}, used in negotiation dialogue systems.


In summary, our contributions are three-fold: (i) we point out human limitations in negotiation and systematically summarize the existing AI solutions aiming to address those limitations; (ii)~we systematically categorize current negotiation 
dialogue benchmarks from a distributive and integrative perspective, and provide an overview of evaluation methods; (iii)~we point out current limitations and promising future research directions.

\section{Negotiations from a Social Science Perspective}
In this section, we will first introduce a framework for human negotiation from social sciences, then discuss 
{human limitations}
in negotiation, which motivates NLP researchers/practitioners to develop strong negotiation dialogue systems. 



\subsection{Understanding of Human Negotiations}
\begin{figure}
    \centering
    \includegraphics[width=0.49\textwidth]{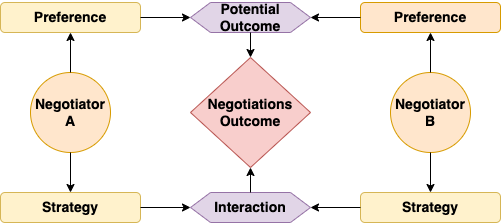}
    \caption{Negotiation Framework for two negotiator scenario from~\citet{brett2016negotiation}.}
    \label{fig:negotationframework}
\end{figure}

\citet{brett2016negotiation} propose a comprehensive framework for a two-party negotiation process, as shown in Figure~\ref{fig:negotationframework}.
Preferences and strategies of the negotiators determine the potential outcomes and the interaction of the negotiation process. 
The preferences of both negotiators create the potential outcome that may be reached by them. The negotiators’ strategies, defined as the goal-directed behaviors that are used in order to reach an agreement~\citep{weingart1990tactical}, affect the interaction, ultimately determining how much of that potential outcome created by the negotiators' preferences is obtained.

\subsection{Human limitations in Negotiation}

Although negotiations are commonly found in daily life (e.g., price bargaining), it is still a challenging task. Without professional training, people often lack the negotiation skills to achieve their desirable goals. They may not know what \emph{strategies} to be used and how to implement these \emph{strategies}. It is also challenging to identify and process implicit information about other negotiators’ interests and preferences in the negotiation. Often times, people view negotiation as a competition and may not even be motivated to seek or express this information ~\citep{brett2016negotiation}. Finally, human cognitive heuristics, biases and emotionality may prove a hindrance in negotiation scenarios. For example, people view themselves, the world and the future as being more positive than in reality~\citep{taylor1989positive}, which may lead to overestimation and optimism in negotiations~\citep{crocker1982biased}. The negotiation could also lead participants to be emotionally engaged and make it more difficult to process information rationally~\citep{pinkley1994conflict}. 
Thus, developing effective negotiation conversational dialogue agents can be beneficial for understanding and controlling for these various factors, and optimizing the negotiation.

\section{Methodology Overviews}
\begin{figure}
    \centering
    \includegraphics[width=0.49\textwidth]{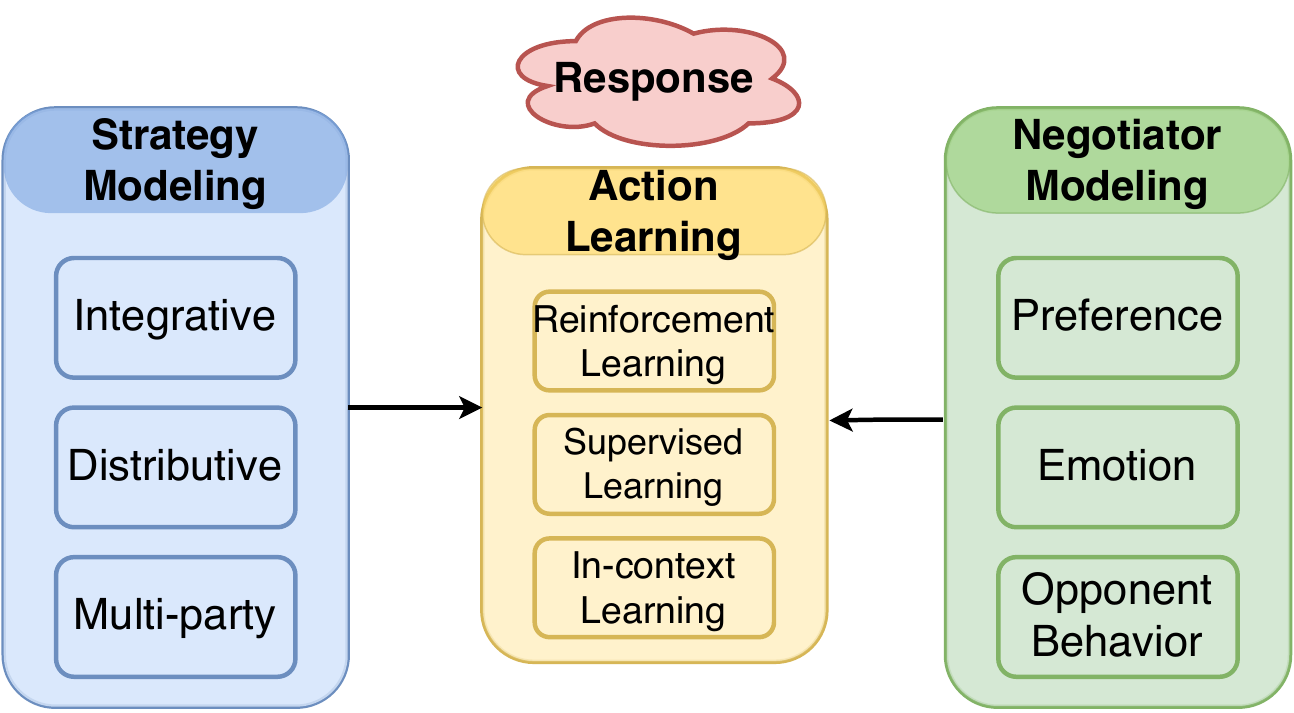}
    \caption{An overview architecture of method section. The \emph{strategy} and \emph{negotiator} modules collect information from the negotiation dialogue, and the \emph{action learning} module conditions on the information and produce responses to push the negotiation forward.}
    \label{fig:method}
\end{figure}

In negotiation dialogues, negotiators interact with each other in a strategic discussion to reach a final goal. As discussed above, \emph{strategies} and \emph{preferences} significantly affect the negotiation outcomes. To effectively assist people in this process, as shown in Figure~\ref{fig:method}, existing research on negotiation dialogues can be categorized into \emph{a)} \emph{Negotiator Modeling}; \emph{b)} 
\emph{Strategy Modeling}; \emph{c)} \emph{Action Learning}.
Herein, \emph{Negotiator Modeling} aims to infer the \emph{explicit information} from other negotiators based on a dialogue context. \emph{Strategy Modeling} learns to select strategies to use given the current dialogue context. Finally, the \emph{Action Learning} incorporates the above negotiation information to map strategies into observable actions or responses, e.g. utterances, by developing dialogue models within the existing machine learning frameworks.




\subsection{Problem Formulation}
Formally, a negotiation dialogue process can be formally characterized as a tuple $({n}, \mathcal{K}, \mathcal{S}, \mathcal{U}, \pi, {g})$. 
Herein, 
{$n$ refers to the number of negotiation party ($n \geq 2$), }
$\mathcal{K}$ refers to the background information for a negotiation dialogue, such as negotiator's preferences and demands towards items. 
This information may not be transparent to others in a dialogue.
$\mathcal{S}$ denotes a strategy trajectory $\{s_1, s_2, ...\}$ used during the negotiation process. 
$\mathcal{U} = \{u_1, u_2, ...\}$ is a sequence of dialogue utterances or actions in a negotiation process. 
A policy $\pi_{\theta}(\mathcal{K}, \mathcal{S}, \mathcal{U})$ is a distribution of actions or a mapping to determine which actions or utterances to produce in order to reach the final negotiation goal $g$.

\subsection{Strategy Modeling}

In negotiations, people use a wide range of tactics and approaches to achieve their goals $g$. Many previous research efforts have focused on modeling these strategies $\mathcal{S}$. They 
can be categorized into three aspects: \emph{integrative} (win-win), such as maximizing unilateral interests~\cite{bazerman1993negotiating}, and \emph{distributive} (win-lost), such as bargaining~\cite{fershtman1990importance}, and \emph{multi-party}~\cite{li2021dadgraph}. 

\subsubsection{Integrative Strategy}
Integrative strategy (known as \emph{win-win}) modeling aims to achieve mutual gains among participants. For instance, \citet{zhao2019rethinking} propose to model the discourse-level strategy using a latent action reinforcement learning (LaRL) framework. LaRL can model strategy transition within a latent space. However, due to the lack of explicit strategy labels, LaRL can only analyze strategies in implicit space. To resolve this problem, \citet{chawla2021casino} define a series of explicit strategies such as \emph{Elicit-Preference}, \emph{Coordination} and \textit{Empathy}. While \textit{Elicit-Preference} is a strategy attempting to discover the preference of an opponent, \textit{Coordination}  promotes mutual benefits through an explicit offer or implicit suggestion. 
In order to capture user's preference, \citet{chawla2022opponent} utilize those strategies using a hierarchical neural model. \citet{yamaguchi-etal-2021-dialogue} also present another collaborative strategy set to negotiate workload and salaries during the interview, whose goal is to reach an agreement between an employer and employee, recommending, for example, to communicate politely, address concerns, and provide side offers.


\subsubsection{Distributive Strategy}
Distributive strategy (known as \emph{win-loss}) modeling focuses on achieving one's own goals and maximizing unilateral interests over mutual benefits. Distributive strategy is used when one insists on their own position or resists the opponent's deal~\cite{zhou-etal-2019-dynamic}. For example, 
to persuade others to donate to a charity, \citet{wang-etal-2019-persuasion} propose a set of persuasion strategies containing 10 different strategies, including logical appeal, emotional appeal, source-related inquiry and others. 
Further exploration on the role of structure (e.g., facing act, emotion)~\cite{li-etal-2020-exploring-role,dutt2020keeping} helps utilize strategy modeling between asymmetrical roles.
Another line of research focuses on the adversarial attack strategy. \citet{dutt-etal-2021-resper} investigate four resisting categories, namely contesting, empowerment, biased processing, and avoidance \cite{fransen2015strategies}. Each individual category contains fine-grained strategic behaviors. For example, contesting refers to attacking the message source, and empowerment implies reinforcing personal preference to contradict a claim (\textit{Attitude Bolstering}) or attempting to arouse guilt in the opponent (\textit{Self Pity}). 

\subsubsection{Multi-party Strategy}

While the previously mentioned work on integrative and distributive strategy modeling mainly relates to two-party negotiations, multi-party strategy modeling is slightly different.  In multi-party situations, strategy modeling needs to consider different attitudes and complex relationships among individual participants, whole groups, and subgroups~\cite{traum2008multi}. \citet{georgila2014single} attempt to model multi-party negotiation using a multi-agent RL framework. Furthermore, \citet{shi2019deep} propose to construct a discourse dependency tree to predict relation dependency among multi-parties. \citet{li2021dadgraph} disclose relations between multi-parties using a graph neural network. However, research in multi-party strategies is currently hindered by limited relevant datasets and benchmarks.

\subsection{Negotiator Modeling}


Negotiation dialogues are affected by various features of negotiators. There is psychological evidence showing that, for example, a negotiation process is affected by personality~\cite{sharma2013role}, relationships~\cite{olekalns2003testing}, 
social status~\cite{blader2012differentiating} and cultural background~\cite{leung2011within}. 
We thus summarize the existing works on modeling negotiators from following three perspectives: \textit{Preference}, \textit{Emotion}, and \textit{Opponent Behavior}. 


\subsubsection{Preference Modeling}
Preference estimation helps an agent infer the intention of their opponents and guess how their own utterances would affect the opponents' preference. 
\citet{nazari2015opponent} propose a simple heuristic frequency-based method to estimate the negotiator's preference. 
However, a critical challenge for preference modeling in negotiation is that it usually requires complete dialogues, so it is difficult to predict those preferences precisely from a partial dialogue.
Therefore, \citet{langlet2018detecting} consider a rule-based system to carefully analyze linguistic features from partial dialogue to identify user's preference.  In further, to enhance preference modeling in those partial dialogues, which widely exist in real-world applications, \citet{chawla2022opponent} formulate preference estimation as a ranking task and propose a transformer-based model that can be trained directly on partial dialogues.

\subsubsection{Emotion Modeling}

Emotion modeling refers to recognizing emotions or emotional changes of negotiators. Explicit modeling of emotions throughout a conversation is crucial to capture and estimate reactions from opponents. To study emotional feelings and expressions in negotiation dialogues, \citet{chawla2021towards} explore the prediction of two important subjective goals, including outcome satisfaction and partner perception. \citet{DBLP:conf/acl/LiuZDSLYJH20} provide explicit modeling on emotion transition engaged using pre-trained language models (e.g., DialoGPT), to support patients. Further, \citet{dutt2020keeping} propose a novel set of dialogue acts modeling \emph{face}, which refers to the public self-image of an individual, in persuasive discussion scenarios. \citet{mishra2022pepds} utilize a reinforcement learning framework to elicit emotions in persuasive messages. 

\subsubsection{Opponent Behavior Modeling}
Opponent behavior modeling refers to detecting and predicting opponents' behaviors during a negotiation process. For example, fine-grained dialogue act labels are provided in the Craigslist dataset~\cite{he2018decoupling}, to help track the behaviors of buyers and sellers.  Based on this information, \citet{zhang2020learningbebavior} propose an opposite behavior modeling framework to estimate opposite action using DQN-based policy learning.
\citet{tran2022ask} leverage dialogue acts to identify optimal strategies for persuading people to donate. \citet{he2018decoupling} firstly propose a framework to decouple the 
opponent behavior modeling with utterance generation, which allows negotiation systems to manage opponent modeling in a precise manner. 
\citet{yang2021improving} further improve the negotiation system with a first-order model based on the theory of Mind~\cite{frith2005theory}, which allows agents to compute an expected value for each mental state. They provided two variants of ToM-based dialogue agents: explicit and implicit, which can fit both pipeline and end-to-end systems.

\subsection{Action Learning}

Action learning empowers negotiation dialogue systems to properly incorporate previous strategies and other negotiator information to generate high-quality responses. Existing research on policy learning can be broadly categorized into \emph{reinforcement learning}, \emph{supervised learning} and \emph{in-context learning}.


\subsubsection{Reinforcement Learning}

\citet{english2005learning} pioneer applying reinforcement learning (RL) techniques to negotiation dialogue systems. 
They propose a single-agent RL framework that learns the policy of two participants individually. However, the single-agent framework is not feasible for situations where two agents interact frequently in a continuously changing environment. \citet{georgila2014single} further propose to use multi-agent RL techniques and provide a way to deal with multi-issue negotiation scenarios. Furthermore, \citet{keizer-etal-2017-evaluating} propose to learn about the actions of negotiators with a Q-learning reward function. They use a Random Forest model trained on a large human negotiation corpus from~\cite{afantenos2012modelling}.


Most recent works have tried to build negotiation dialogue models using RL techniques with deep learning. \citet{zhang2020learningbebavior} propose OPPA, which utilizes the system actions to estimate how a target agent behaves. The system actions are predicted based on the target agent's actions. The reward of the executed actions is obtained by predicting a structured output given a whole dialogue.
Additionally, \citet{shi2021refine} use a modular framework containing a language model to generate responses. A response detector would automatically annotate the response with a negotiation strategy and an RL-based reward function to assign a score to the strategy. 
However, this modular framework separates policy learning from response generation. \citet{gao2021deepRL} propose an integrated framework with deep Q-learning, which includes multiple channel negotiation skills. It allows agents to leverage parameterized DQN to learn a comprehensive negotiation strategy that integrates linguistic communication skills and bidding strategies.

\subsubsection{Supervised Learning}
 
Supervised learning (SL) is another popular paradigm for policy learning. \citet{lewis2017deal} adopt a Seq2Seq model to learn what action should be taken by maximizing the likelihood of the training data. However, supervised learning only aims to mimic the average human behavior, so~\citet{he2018decoupling} propose to apply a supervised model to directly optimize a particular dialogue reward function, which is characterized by i) the utility function of the final price for the buyer and seller ii) the differences between two agents’ utilities iii) the number of utterances in the dialogue. \citet{zhou2019augmenting} first train a strategy predictor to predict whether a certain negotiation strategy occurred in the next utterance using supervised training. Then, the response generation conditions on the predicted negotiation strategy, as well as user utterance and dialogue context. In addition, \citet{joshi2021dialograph} incorporate a pragmatic strategies graph network with the seq2seq model to create an interpretable policy learning paradigm.  Recently, \citet{dutt2021resper} propose a generalized
framework for identifying resisting strategies in persuasive negotiations using a pre-trained BERT model~\cite{devlin2019bert}. In addition, there are also research attempts to jointly train several sub-tasks simultaneously.
\citet{li2020end} propose an end-to-end framework that integrates several sub-tasks, including intent and semantic slot classification, response generation and filtering tasks in a Transformer-based pre-trained model. \citet{zhou2019augmenting} propose jointly modelling semantic and strategy history using finite state transducers (FSTs) with hierarchical neural models. \citet{chawla2022opponent} integrate a preference-guided response generation model with a ranking module to identify opponents' priority.

\subsubsection{In-context Learning}
With the recent emergence of large language models such as GPT-3.5 and GPT-4\footnote{https://platform.openai.com/docs/models/}, a few studies have applied zero-shot and few-shot in-context learning. These techniques leverage the inherent knowledge of LLMs to predict agent behaviors and generate utterances. \citet{fu2023improving} utilize LLMs in the context of bargaining, while \citet{xu2023exploring} employ them for the popular game ``Werewolf''. Besides, \citet{chen2023put} propose a framework to evaluate strategic planning and execution of LLM agents. In both tasks, the LLMs act as agents, negotiating with other LLMs under specific scenarios to achieve pre-defined goals.





\section{Negotiation Datasets}
\label{datasetsection}
\begin{table*}[t]
\begin{center}
\tiny
\setlength{\tabcolsep}{1.6mm}{
\begin{tabular}{lcccccc}
\toprule
DataSet & Negotiation Type & Scenario & \# Dialogue  & \# Avg. Turns & \# Party & \# Modality \\
\midrule
InitiativeTaking (\citet{nouri-traum-2014-initiative}) & Integrative & Fruit Assignment & 41  & - & Multi & -\\
STAC (\citet{stac2016discourse}) & Integrative & Strategy Games & 1081 & 8.5  & Two & - \\
DealorNoDeal (\citet{lewis2017deal}) & Integrative & Item Assignment & 5808 & 6.6 & Two & - \\
Craigslist (\citet{he2018decoupling}) & Distributive & Price Bargain & 6682 & 9.2 & Two & - \\
{M3} (\citet{kontogiorgos2018multimodal}) & Integrative & Object Moving &  15 & - & Multi & MultiModal \\
{Niki \& Julie} (\citet{artstein2018niki}) & Integrative & Item Ranking & 600 & - & Two & MultiModal \\
NegoCoach (\citet{zhou-etal-2019-dynamic}) & Distributive & Price Bargain & 300 & - & Two & - \\
PersuasionforGood (\citet{wang-etal-2019-persuasion}) & Distributive & Donation & 1017 & 10.43 & Two & -\\
FaceAct (\citet{dutt2020keeping}) & Distributive & Donation & 299 & 35.8 & Two & - \\
AntiScam (\citet{li2020end}) & Distributive & Privacy Protection & 220  & 12.45 & Two & - \\
CaSiNo (\citet{chawla2021casino}) & Integrative & Item Assignment & 1030 & 11.6 & Two & - \\
JobInterview (\citet{yamaguchi-etal-2021-dialogue}) & Integrative & Job Interview & 2639 & 12.7 & Two & - \\
{DeliData} (\citet{karadzhov2021delidata}) & Integrative & Puzzle Game & 500 & 28 & Multi & - \\
DinG (\citet{boritchev-amblard-2022-multi}) & Integrative & Strategy Game & 10 & 2357.5 & Multi & -\\
NegoBar (\citet{heddaya2023language}) & Distributive & Price Bargain & 408 & 35.85 & Two & - \\
\bottomrule
\end{tabular}}
\caption{Negotiation dialogues benchmarks are sorted by their publication time. For each dataset, we present the negotiation type, scenario, the number of dialogues and corresponding average turns, and party attributes.}
\label{tab:dataset}
\end{center}
\end{table*}
In this section, we summarize the existing negotiation datasets and resources. 
Table~\ref{tab:dataset} shows all of the 14 collected benchmarks, along with their negotiation types, scenarios, data scale and modality. We categorize these benchmarks based on their negotiation types, namely, \textit{integrative} negotiation and \textit{distributive} negotiation. 

\subsection{Integrative Negotiation Datasets}
In integrative negotiations, there is normally more than one issue being negotiated. To achieve optimal negotiation goals, the involved players should make trade-offs for these multiple issues.

\paragraph{Multi-player Strategy Games} Strategy video games provide ideal platforms for people to verbally communicate with other players to accomplish their missions and goals.~\citet{stac2016discourse} propose the STAC benchmark, which is based on the game of Catan. In this game, players need to gather resources, including wood, wheat, sheep, and more, with each other to purchase settlements, roads and cities. As each player only has access to their own resources, they have to communicate with each other. To investigate the linguistic strategies used in this situation, STAC also includes an SDRT-styled discourse structure. \citet{boritchev-amblard-2022-multi} also collect a \emph{DinG} dataset  from French-speaking players in this game. The participants are instructed to focus on the game, rather than talk about themselves. As a result, the collected dialogues can better reflect the negotiation strategy used in the game process.

\paragraph{Negotiation for Item Assignment}
Item assignment scenarios involve a fixed set of items as well as a predefined priority for each player in the dialogue. As the players only have access to their own priority, they need to negotiate with each other to exchange the items they prefer. \citet{nouri-traum-2014-initiative} propose \emph{InitiativeTalking}, occurring between the owners of two restaurants. They discuss how to distribute the fruits (i.e., apples, bananas, and strawberries) and try to reach an agreement. \citet{lewis2017deal} propose \emph{DealorNoDeal}, a similar two-party negotiation dialogue benchmark where both participants are only shown their own sets of items with a value for each and both of them are asked to maximize their total score after negotiation. \citet{chawla2021casino} propose \emph{CaSiNo}, a dataset on campsite scenarios involving campsite neighbors negotiating for additional food, water, and firewood packages. Both parties have different priorities over different items.

\paragraph{Negotiation for Job Interview}
Another commonly encountered negotiation scenario is job offer negotiation with recruiters. \citet{yamaguchi-etal-2021-dialogue} fill this gap and propose the \emph{JobInterview} dataset. \emph{JobInterview} includes recruiter-applicant interactions over salary, day off, position, and workplace. Participants are informed with opposite’s preferences and the corresponding issues. Feedback from the opposites will be forwarded to participants during the negotiation process.

\subsection{Distributive Negotiation Datasets}
Distributive negotiation is a discussion over a fixed amount of value (i.e., slicing up the pie). In such negotiation, the involved people normally talk about a single issue (e.g., item price) and therefore, there are hardly trade-offs between multiple issues in such a negotiation. 

\paragraph{Persuasion For Donation}
Persuasion, convincing others to take specific actions, is a necessary required skill for negotiation dialogue~\cite{sycara1990persuasive,sierra1997framework}. \citet{wang-etal-2019-persuasion} focus on persuasion and propose \emph{PersuasionforGood}, two-party persuasion conversations about charity donations. In the data annotation process, the persuaders are provided some persuasion tips and example sentences, while the persuaders are only told that this conversation is about charity. The annotators are required to complete at least ten utterances in a dialogue and are encouraged to reach an agreement at the end of the conversations. \citet{dutt2020keeping} further extend \emph{PersuasionforGood} by adding the utterance-level annotations that change the positive and/or the negative face acts of the participants in a conversation.
A face act can either raise or attack the positive or negative face of opponents in the conversation.

\paragraph{Negotiation For Product Price}
Negotiations over product prices can be observed on a daily basis. 
\citet{he2018decoupling} propose \emph{CraigslistBargain}, a negotiation benchmark based on a realistic item price bargaining scenario. In \emph{CraigslistBargain}, two agents, a buyer and a seller, are required to negotiate the price of a given item. The listing price is available to both sides, but the buyer has a private price. Two agents chat freely to decide on a final price. The conversation is completed when both agents agree on a price or one of the agents quits. \citet{zhou-etal-2019-dynamic} propose \emph{NegoCoach} benchmark on similar scenarios, but with an additional negotiation coach who monitors messages between the two annotators and recommends tactics in real-time to the seller to get a better deal.

\paragraph{User Privacy Protection}
Privacy protection of negotiators has become more and more vital. Participant (e.g., attackers and defenders) goals are also conflicting. \citet{li2020end} propose \emph{Anti-Scam} benchmark which focuses on online customer service. In \emph{Anti-Scam}, users try to defend themselves by identifying whether their components are attackers who try to steal sensitive personal information. \emph{Anti-Scam} provides an opportunity to study human elicitation strategies in this scenario. 
\section{Evaluation}

We categorize the methods for evaluating the negotiation dialogue systems into three types: \emph{goal-oriented} evaluation, \emph{game-based} evaluation and \emph{human} evaluation.
Table~\ref{tab:metrics} summarizes the evaluation metrics that are introduced in our survey. 

\subsection{Goal-based Metrics}
Goal-oriented metrics mainly refer to the quantifiable measures on evaluating agent's proximity to the negotiation goals from the perspective of strategy modeling,
task fulfillment,
and sentence realization. 
\textit{Success Rate (SR)}~\cite{zhao2019rethinking} is the most widely used metric to measure how frequently an agent completes the task within their goals.
Meanwhile, \textit{Prediction Accuracy (PA)} and \emph{macro/average F1 score} are also employed to evaluate the accuracy of agent's strategy predictions~\cite{nouri-traum-2014-initiative, wang-etal-2019-persuasion,dutt2020keeping,chawla2021casino}.
Specifically, \citet{yamaguchi-etal-2021-dialogue} present a task where the model is required to label the human-human negotiation outcomes as either a success or a breakdown, and use following metrics: \emph{area under the curve} (ROC-AUC), \emph{confusion matrix} (CM), and \emph{average precision} (AP) to evaluate the model.
Moreover, \citet{DBLP:conf/naacl/KornilovaED22} introduce Item Response Theory (IRT) to analyze the effectiveness of persuasion on the audience.

In terms of language realization for negotiation dialogue,
\citet{DBLP:books/sp/15/HiraokaNSTN15} employ a pre-defined naturalness metric 
(i.g., a bi-gram overlap between the prediction and ground-truth) 
as part of the reward to evaluate policies in negotiation dialogues.
Other classical metrics for evaluating the quality of response are also used, i.e., perplexity (PPL), BLEU-2, ROUGE-L, and BOW Embedding-based Extrema matching score~\cite{lewis2017deal}.

\begin{table}[!t]
\begin{adjustbox}{ width=0.45\textwidth,center}
\centering
\tiny
\begin{tabular}{lc}
\toprule
    \makecell[c]{Goal-based \\ Metrics}  & \makecell[c]{SR (\citeyear{zhao2019rethinking}); PA (\citeyear{nouri-traum-2014-initiative,wang-etal-2019-persuasion,dutt2020keeping}); Average F1 score (\citeyear{chawla2021casino}); \\ Macro F1 score (\citeyear{wang-etal-2019-persuasion,dutt2020keeping}); ROC-AUC, CM, AP (\citeyear{yamaguchi-etal-2021-dialogue}); IRT (\citeyear{DBLP:conf/naacl/KornilovaED22}); \\ Naturalness (\citeyear{DBLP:books/sp/15/HiraokaNSTN15});  PPL, BLEU-2, ROUGE-L, Extrema (\citeyear{lewis2017deal})} \\ 
\midrule
    \makecell[c]{Game-based \\ Metrics} & \makecell[c]{WinRate, AvgVPs (\citeyear{keizer-etal-2017-evaluating}); Utility, Fairness, Length (\citeyear{he2018decoupling});\\ Avg. Sale-to-list Ratio, Task Completion Rate (\citeyear{zhou-etal-2019-dynamic}); Robustness (\citeyear{cheng2019evaluating})} \\ 
\midrule
    \makecell[c]{Human \\ Evaluation} & \makecell[c]{Customer satisfaction, Purchase decision, Correct response rate (\citeyear{DBLP:books/sp/15/HiraokaNSTN15}); \\ Achieved agreement rate, Pareto optimality rate (\citeyear{lewis2017deal}); Likert score (\citeyear{he2018decoupling})} \\ 
\bottomrule
\end{tabular}
\end{adjustbox}
\caption{Various Metrics used in the existing negotiation dialogues benchmarks.}
\vspace{-2mm}
\label{tab:metrics}
\end{table}

\subsection{Game-based Metrics}
Different from the goal-oriented metrics that focus on measuring how successful an agent achieves the negotiation goals, game-based evaluation provides a user-centric perspective to evaluate systems. \citet{keizer-etal-2017-evaluating} measure agent's ability on negotiation strategy prediction within the online game ``\emph{Settlers of Catan}''. They propose the metrics \emph{WinRate} 
and \emph{AvgVPs} to evaluate the success of human and agent seperately. 
\citet{he2018decoupling} present a task where two agents bargain to get the best deal using natural language. They use task-specific scores to test the performance of the agents, including: \emph{utility},
\emph{fairness},
and \emph{length}.
\citet{zhou-etal-2019-dynamic} design a task where a seller and a buyer try to achieve a mutually acceptable price through a natural language negotiation. They adopt \emph{average sale-to-list ratio} and \emph{task completion rate} to evaluate  agent performance. 
Besides, \citet{cheng2019evaluating} propose an adversarial attacking evaluation approach to test the \emph{robustness} of negotiation systems.

\subsection{Human Evaluation}
To evaluate the users' satisfaction with the dialogue systems, human judgment is employed as a subjective evaluation of agent performance.
\citet{DBLP:books/sp/15/HiraokaNSTN15} use a user simulator as the salesperson to bargain with customers in real and have the users annotate subjective \emph{customer satisfaction} (a five-level score), the final decision of making a purchase (a binary number indicating whether persuasion is successful), and the \emph{correct response rate} in the dialogues.
\citet{lewis2017deal} employ crowd-sourcing workers to highlight that essential information when bargaining with negotiation systems, covering the percentage of dialogues where both interlocutors finally achieve an agreement, and \emph{Pareto optimality}, i.e., the percentage of the Pareto optimal solutions in all the agreed deals. 
\citet{he2018decoupling} propose human likeness as a metric in evaluating how well the dialogue system is doing in a bargain. They ask workers to manually score the dialogue agent using a \emph{Likert} metric to judge whether the agent acts like a real human or not.


\section{New Frontiers and Challenges}
The previous sections summarize the prominent achievements of previous work in negotiation dialogue, including benchmarks, evaluation metrics, and methodology. In this section, we will discuss some new frontiers that allow negotiation dialogue systems to be fit to actual application needs and to be applied in real-world scenarios.

\paragraph{Multi-modal Negotiation Dialogue}
Existing research works in negotiation dialogue rarely consider multi-modality. However, humans tend to perceive the world in multi-modal patterns, not limited to text but also including audio and visual information. For example, the facial expressions and emotions of participants in a negotiation dialogue could be important cues for making negotiation decisions. Further work can consider adding this non-text-based information into the negotiation.

\paragraph{Multi-Party Negotiation Dialogue}
Although some work sheds light on multi-party negotiation, most current negotiation dialogue benchmarks and methods predominantly focus on two-party settings. Therefore, multi-party negotiation dialogues are underexplored. Future work can consider collecting dialogues in multi-party negotiation scenarios, including \emph{General multi-party negotiation} and \emph{Team negotiation.} Specifically, \emph{General multi-party negotiation} is a type of bargaining where more than two parties negotiate toward an agreement. For example, next-year budget discussion with multiple department leaders in a large company. \emph{Team negotiation} is a team of people with different relationships and roles. It is normally associated with large business deals and highlights the significance of relationships between multi-parties. There could be several roles, including leader, recorder, and examiner, in a negotiation team~\cite{halevy2008team}.

\paragraph{Cross-Culture \& Multi-lingual Negotiation Dialogue}
Existing negotiation dialogue benchmarks overwhelmingly focus on English while leaving other languages and cultures under-explored. With the acceleration of globalization, a dialogue involving individuals from different cultural backgrounds~\cite{chawla2023social,zhan2023socialdial,joshi2024natural} becomes increasingly important and necessary. There is an urgent need to provide people with a negotiation dialogue system that is multicultural and multi-lingual. Further works can consider incorporating multi-lingual utterances and social norms among different countries into negotiation dialogue benchmarks.


\paragraph{Negotiation Dialogue in Real-world Scenarios}
As discussed in Section~\ref{datasetsection}, previous works have already proposed many negotiation dialogue benchmarks in various scenarios. However, we notice that most of these benchmarks are created through human crowd-sourcing. Participants are often invited to play specific roles in the negotiation dialogue. The resulting dialogues may not perfectly reflect the negotiations in real-world scenarios (e.g., politics, business). Therefore, it could be a promising research direction to collect real-world negotiation dialogues. For example, one could collect recorded business meetings or phone calls.


\section{Conclusion}

This paper presents the first systematic review on the progress of negotiation dialogue systems. We firstly provide an understanding of  negotiation between humans from a social science perspective.
Then we thoroughly summarize the existing works, which covers various domains and highlight their challenges, respectively. We additionally summarize currently available methodologies, benchmarks, and evaluation methods. We also shed light on some new trends in this research field. We hope this survey inspires and facilitates future research on negotiation dialogue systems.

\section*{Limitations}

This survey  briefly introduced the motivation and limitation of human negotiation from social science perspectives, and summarized methodology, dataset and evaluation methods in the field of computational linguistics. The limitation relays on that we only have brief investigation on the human negotiation. Further, we will conduct a comprehensive investigation from the social science perspectives and then motivate our future work in the dialogue research. In further, we will summarize the details of each paper and illustrate the difference between these papers. Nevertheless, we hope that our survey will inspire and facilitate future research as a good foundation.

\section*{Acknowledgements}

We would like to thank the anonymous reviewers for their valuable comments and suggestions. This material is based on research sponsored by DARPA under
agreement number HR001122C0029. The U.S. Government is authorized to reproduce and distribute reprints for Governmental purposes notwithstanding any copyright notation thereon.


\bibliography{ref_rebiber,anthology}
\bibliographystyle{acl_natbib}




\end{document}